\documentclass{article}
\usepackage{spconf,amsmath,graphicx}

\usepackage{enumitem}
\setlist{nosep, leftmargin=14pt}

\usepackage{mwe} 

\usepackage{lmodern,bm}
\usepackage{soul}
\sethlcolor{red}

\usepackage{amssymb}

\usepackage{caption}

\graphicspath{ {./} }

\title{Few-shot adaptation for morphology-independent cell instance segmentation}

\name{Ram J. Zaveri$^{\star}$\thanks{$^{\star}$Denotes equal contribution.} \qquad Voke Brume$^{\star}$ \qquad Gianfranco Doretto}

\address{Lane Department of Computer Science and Electrical Engineering\\West Virginia University, Morgantown, WV 26506 USA}

\begin{document}
%
\maketitle
\begin{abstract}
Microscopy data collections are becoming larger and more frequent. Accurate and precise quantitative analysis tools like cell instance segmentation are necessary to benefit from them. This is challenging due to the variability in the data, which requires retraining the segmentation model to maintain high accuracy on new collections. This is needed especially for segmenting cells with elongated and non-convex morphology like bacteria. We propose to reduce the amount of annotation and computing power needed for retraining the model by introducing a few-shot domain adaptation approach that requires annotating only one to five cells of the new data to process and that quickly adapts the model to maintain high accuracy. Our results show a significant boost in accuracy after adaptation to very challenging bacteria datasets.
\end{abstract}

\section{Introduction}
\label{sec:intro}

The increased frequency and size of microscopy data collections enable rapid  scientific discovery but require improved automated tools for quantitative analysis~\cite{Jeckel2021-cz}. Image collections present a remarkable variability, depicting cells, membranes, and organelles at various sizes and shapes, collected with different microscope configurations, and from tissues treated differently. In particular, images of bacteria cells present additional challenges due to their smaller size and elongated and non-convex morphologies, including shapes with multiple branches~\cite{Ducret2016-hm}.

The typical approach to tasks such as automated cell instance segmentation on these collections is to rely on a method trained on a large \emph{source} dataset~\cite{Stringer2021-yw,Cutler2022-js}. However, because of their variability, new \emph{target} image collections will very likely have different statistical properties than those used for training the segmentation method. Because of this \emph{distribution shift}~\cite{Shimodaira2000-rp}, the segmentation accuracy will deteriorate on new collections. The obvious solution is to retrain the segmentation model on the target data. This is effective for methods like~\cite{schmidt2018,Greenwald2021-uj,Stringer2021-yw} for handling more regular, nearly-convex cell morphologies, whereas for more difficult morphologies (e.g., for bacteria), methods like~\cite{Panigrahi2021-fc,Cutler2022-js} will be more appropriate. The problem is that this approach requires a labor-intensive data annotation process and a computationally expensive training procedure.

A different solution is to use a \emph{domain adaptation} approach~\cite{Motiian2017-ow}, where a portion of the new \emph{target} data is used for ``adapting'' the model to perform well in spite of the distribution shift. Current approaches are either modality-dependent and not tailored to cell segmentation~\cite{Guan2022-qj,Xun2021-ne,Goetz2016-az,Zhu2020-mv,Bermudez-Chacon2018-bf} or computationally demanding because they are unsupervised and need to process large amounts of data~\cite{Javanmardi2018-ym,Zhang2020-eu,Yang2021-qs,Liu2021-nw}. Recently, CellTranspose, a few-shot domain adaptation approach for cell instance segmentation, was proposed~\cite{Keaton2023}. It has the advantage of requiring only a few cells (e.g., one to five) to be annotated by the user. Then, the model is quickly adapted to the target data domain with minimal computational cost. Despite offering an appealing solution, CellTranspose's ability to handle elongated and non-convex morphologies is limited by the adopted segmentation strategy. This is based on predicting segmentation masks along with gradient flows converging towards the cell centers~\cite{Stringer2021-yw}. This approach has been shown to have limitations with morphologies that are not nearly-convex and for which a global cell center cannot be correctly defined~\cite{Cutler2022-js}.

In this work, we propose a few-shot domain adaptation approach for cell instance segmentation that addresses the limitations expressed above. We do so by starting from a segmentation strategy that handles elongated and non-convex morphologies by predicting segmentation boundaries and a boundary distance field along with its gradient. This requires designing new training losses to support the fast adaptation of a pretrained model to the target domain with just a few (one to five) annotated cells. The results show that our framework effectively supports adaptation to multiple datasets in the case of challenging non-convex morphologies.


\section{Method}
\label{sec:method}

We are interested in segmenting instances of cells, like bacteria, which can have small dimensions (e.g., a hundred of pixels in area), and can exhibit a wide range of morphologies~\cite{Kysela2016-qf}. Also, new image collections are deeply affected by the different optical features determined by the tissue treatment or microscopy techniques. This means that given new data to segment, a state-of-the-art tool like Omnipose~\cite{Cutler2022-js} will be affected by the distribution shift~\cite{Shimodaira2000-rp}, if not retrained on such new data, and it will significantly underperform.

To avoid the costly process of annotating large amounts of new data and intense network training for every new segmentation task, we propose a two-step approach. First, we rely on a robust and precise segmentation method that can handle a diversity of morphologies, and that has been pre-trained on a \emph{source} dataset $\mathcal{D}^s$. Second, we take a \emph{few-shot} learning approach, whereby given a \emph{target} dataset $\mathcal{D}^t$, distributed differently than $\mathcal{D}^s$, we require the user to label only a \emph{minimal} amount of target data, let us say $K$ image patches for a $K$-shot learning, essentially depicting one cell each. The pre-training step needs to be done only once. The \emph{few-shot adaptation} step is fast and low-cost both in manual labor and compute power, and is easily repeatable for new datasets to process.

Note that mere model fine-tuning on the $K$ image patches is not a viable option since this would lead to overfitting~\cite{Goodfellow2016-iu}. Hence, we introduce a proper few-shot supervised domain adaptation method for cell instance segmentation. This work overcomes the limitations of previous work~\cite{Keaton2023}, which could not handle the adaptation to datasets with a large diversity of morphologies. Thus, to the best of our knowledge, this is the first few-shot adaptation for cell instance segmentation that is morphology-independent, meaning it can handle cases where a cell center cannot be correctly defined, which typically occur with shapes with elongated and/or multiple branches.

\begin{figure}[t!]
\centering
\includegraphics[width=0.5\textwidth]{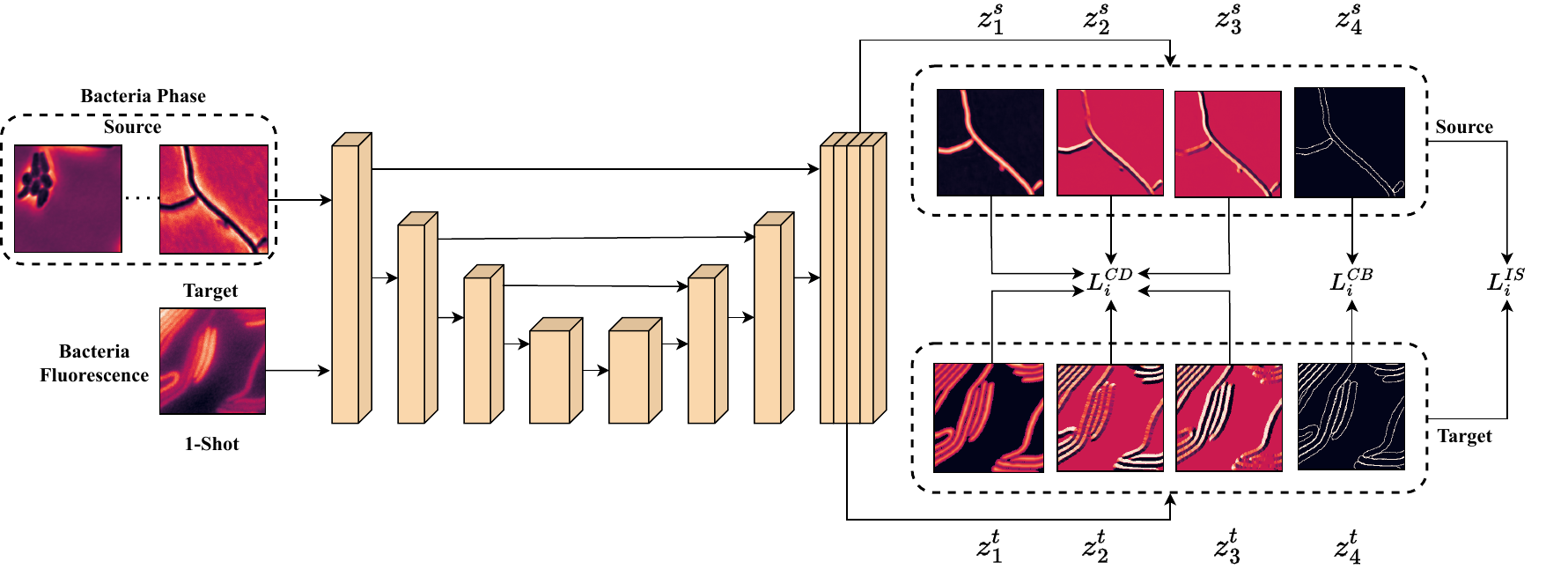}
\caption{\textbf{Architecture approach.} Illustration of our contrastive learning-based few-shot cellular instance segmentation approach.}
\label{fig:adaptation_approach}
\vspace{-5mm}
\end{figure}

\subsection{Pretrained Model}
\label{sec-pretrained-model}

We rely on Omnipose~\cite{Cutler2022-js} as the pretrained method because it is the state-of-the-art for handling morphology-independent cell instance segmentation. We summarize it here to introduce notation. Given an image $I$, a network $f$ produces a dense, pixel-wise feature $\bm{Z} = f(I)$, where $ \bm{Z} = [Z_1, Z_2, Z_3, Z_4] \in \mathbb{R}^{h \times w \times 4}$. For a pixel $i$, the feature $\mathbf{z} = [z_1, z_2, z_3, z_4] \in \bm{Z}$, has the following meaning. $z_1 \doteq \phi$, where $\phi$ is the \emph{distance field}, which is the distance between pixel $i$ inside a cell region to the closest point on the cell boundary. $\phi=0$ if pixel $i$ is outside of a cell. Then, $\bm{u} \doteq (z_2, z_3)$ is the  \emph{gradient flow field}, which is the normalized gradient of the distance field, pointing towards the medial axis (skeleton) of the cell that is defined by the stationary points of the distance field. Finally, $z \doteq z_4$ represents the unnormalized score indicating the probability of pixel $i$ belonging to a cell boundary. With this notation, we rewrite the feature as $\mathbf{z} = [\phi, \bm{u}, z]$. The network $f$ is trained in a supervised manner with pixel-wise instance segmentation loss
\begin{equation}
  \small
  \mathcal{L}_i^{IS} = (\phi - \mathtt{d})^2 + \nu \| \bm{u} - (\mathtt{g_x},  \mathtt{g_y}) \|^2 + \mu H(\mathtt{b},\sigma(z))  \; ,
  \label{eq-instance-segmentation}
\end{equation}
where, for pixel $i$, $\mathtt{d}$ is the ground-truth distance field,   $(\mathtt{g_x}, \mathtt{g_y})$ is the ground-truth gradient flow field with unit $\ell_2$-norm, $\mathtt{b} \in \{0, 1\}$ is the binary mask label indicating absence/presence of a cell border, $\sigma(z) \doteq 1/(1+\exp(-z))$, $H$ is the binary cross-entropy. $\nu$ and $\mu$ are hyperparameters set to $0.5$ and $1.0$. For image $I$, the contributions $\{ \mathcal{L}_i^{IS} \}$ are aggregated into the loss $\mathcal{L}^{IS} = \sum \mathcal{L}_i^{IS} + \mathcal{L}_{IVP}$, where $\mathcal{L}_{IVP}$ is a loss term that takes into account the initial value problem satisfied by the distance field~\cite{Cutler2022-js}.

From $\bm{Z}$, a segmentation head $g$ produces the mask $Y = g(\bm{Z})$ based on an Euler integration~\cite{Cutler2022-js}. For pixel $i$, the predicted label is $y \in \{0, 1, \cdots, N \}$. $N$ is the total number of cell instances segmented. $y=0$ indicates absence of a cell. See also Figure~\ref{fig:adaptation_approach}.

\subsection{Adaptation of the Pretrained  Model}

A cell target pixel $i$ of $I \in \mathcal{D}^t$, with label $(\mathtt{d}^t, \mathtt{g_x}^t, \mathtt{g_y}^t, \mathtt{b}^t)$, should have its feature $\mathbf{z}^t$ close to the pixel features from source images in $\mathcal{D}^s$ with the same label. However, this is generally not the case because of the \emph{shift} between source and target domains. Therefore, we design new training losses to reverse the effects of domain shift, which are tailored specifically to the network $f$ introduced in \S\ref{sec-pretrained-model}, and that can adapt $f$ to the target to prevent performance deterioration of the segmentation process.

\textbf{Contrastive Distance Loss. }
We jointly align $\phi^t$ and $\bm{u}^t$ with the distance and its gradient features of source pixels by setting up a contrastive prediction task~\cite{Chen2020-zr}. This differs from~\cite{Keaton2023} where the alignment was needed only for the gradient features. To that end, we identify a \emph{positive} source pixel inside the boundaries defined by $\mathtt{b}^s_+$, with distance $\phi^s_+$ and gradient features $\bm{ u}^s_+$ that best match the label $(\mathtt{d}^t,\mathtt{g_x}^t, \mathtt{g_y}^t)$, according to a similarity measure. We use $s((\phi,\bm{u}), (\psi,\bm{v})) \doteq \exp{[\frac{-1}{2\sigma}(\phi - \psi)^2]} \bm{u}^{\top} \bm{v} / \|\bm{u} \| \| \bm{v} \|$, which combines a radial basis kernel with parameter $\sigma$  to compare the distance features, and a cosine similarity kernel to compare the gradients, where $\|\cdot\|$ denotes $\ell_2$-norm. We also compose the set of \emph{negative} features $\mathcal{N}_i$, taken from inside the boundaries $\mathtt{b}^s_-$, and such that $\mathcal{N}_i = \{ (\phi_-^s,\bm{ u}^s_-) \; | \; s((\phi_+^s,\bm{ u}^s_+), (\bm{ u}^s_-, \phi_-^s)) < \delta \}$, where $\delta$ is a threshold that we choose. This allows to define the following \emph{contrastive distance loss}
\begin{equation}
  \small
  \mathcal{L}_i^{CD} = - \log \frac{e^{ \frac{s ((\phi^t,\bm{ u}^t), (\phi_+^s,\bm{ u}^s_+))}{\tau} }}{ e^{\frac{s ((\phi^t,\bm{ u}^t), (\phi_+^s, \bm{ u}^s_+))}{\tau} } + \!\!\!\! \sum\limits_{{(\phi^s_-, \bm{ u}^s_-) \in \mathcal{N}_i}} \!\!\! e^{\frac{s ((\phi^t,\bm{ u}^t), (\phi^s_-, \bm{ u}^s_- ) )}{\tau} } }
  \label{eq-contrastive-flow-loss}
\end{equation}
where $\tau$ is a temperature parameter. $\mathcal{L}_i^{CD}$ pulls the positive pair $((\phi^t,\bm{ u}^t), (\phi_+^s,\bm{ u}^s_+))$ closer, while pushing apart every negative pair $((\phi^t,\bm{ u}^t), (\phi^s_-, \bm{ u}^s_- ) )$. We also extend the strategy of~\cite{Keaton2023} for mining hard negatives for $\mathcal{N}_i$.

For a target image and a source image pair we aggregate the loss contributions from the target pixels inside the boundaries $\mathtt{b}^t$. If $\mathcal{B}$ indicates this set of pixels, then the aggregate loss is
$\mathcal{L}^{CD} = \frac{1}{| \mathcal{B} |} \sum_{\mathcal{B}} \mathcal{L}^{CD}_i$.

\textbf{Contrastive Boundary Loss. }
We align the unnormalized binary classification score $z^t$ with the scores of the source pixels with same label. We derive a contrastive loss for boundary detection by treating this as a binary classification adaptation problem~\cite{Motiian2017-ow}. This differs from~\cite{Keaton2023} because there they need to focus on cell masks rather than boundaries.

We want to pull together $z^t$ and the unnormalized score $z_+^s$, of pixels in source images with label $\mathtt{b}^s = \mathtt{b}^t$. We also need to minimize the similarity between $z^t$ and the unnormalized score $z_-^s$, of pixels in source images with label $\mathtt{b}^s \ne \mathtt{b}^t$. This is achieved by minimizing this \emph{contrastive boundary loss}
\begin{equation}
  \mathcal{L}^{CB} = \frac{1}{|\mathcal{P}|} \sum_{\mathcal{P}}   d(z^t,z_+^s) +\lambda \frac{1}{|\mathcal{N}|} \sum_{\mathcal{N}}   k(z^t,z_-^s) \; .
  \label{eq-boundary-loss}
\end{equation}
In~\eqref{eq-boundary-loss}, for a target image and a source image pairs, $\mathcal{P}$ is the set of positive pairs $\{(z^t,z_+^s)\}$, and $\mathcal{N}$ is the set of negative pairs $\{(z^t,z_-^s)\}$. In addition, $d(z^t,z_+^s) = \frac{1}{2} (z^t - z_+^s)^2$, $k(z^t,z_-^s) = \frac{1}{2} \max(0,m - | z^t - z_-^s | )^2$, $m$ is a margin, and $\lambda$ is a hyperparameter.

\textbf{Few-shot Adaptation. }
Let $\mathcal{D}_K^t$ indicate the $K$ labeled samples of the target dataset $\mathcal{D}^t$. The $K$\emph{-shot adaptation} training minimizes 
\begin{equation}
  \mathcal{L}^{ISA} = \sum_{\mathcal{D}_K^t} \left( \mathcal{L}^{IS} + \frac{\gamma_1}{|\mathcal{D}^s|} \sum_{\mathcal{D}^s} \mathcal{L}^{CB} + \frac{\gamma_2}{|\mathcal{D}^s|} \sum_{\mathcal{D}^s} \mathcal{L}^{CD} \right) \; .
  \label{eq-adaptation}
\end{equation}
The training assumes that the model has already been pretrained. We used Omnipose pretrained for 3800 epochs, and we adapt the model for 5 epochs according to~\eqref{eq-adaptation}. Source images are continuously randomly paired with one of the $K$ target samples without replacement to ensure that even a 1-shot adaptation performs a significant pull of the model towards the target distribution. After adaptation the model is fine-tuned on $\mathcal{D}_K^t$ with $\mathcal{L}^{IS}$ for 5 more epochs with a very low learning rate of $1e-7$. For the definition of one ``shot'' we followed~\cite{Keaton2023}, where the nominal cell size is computed according to~\cite{Cutler2022-js}.


\begin{figure}[t!]
\centering
\includegraphics[width=0.48\textwidth]{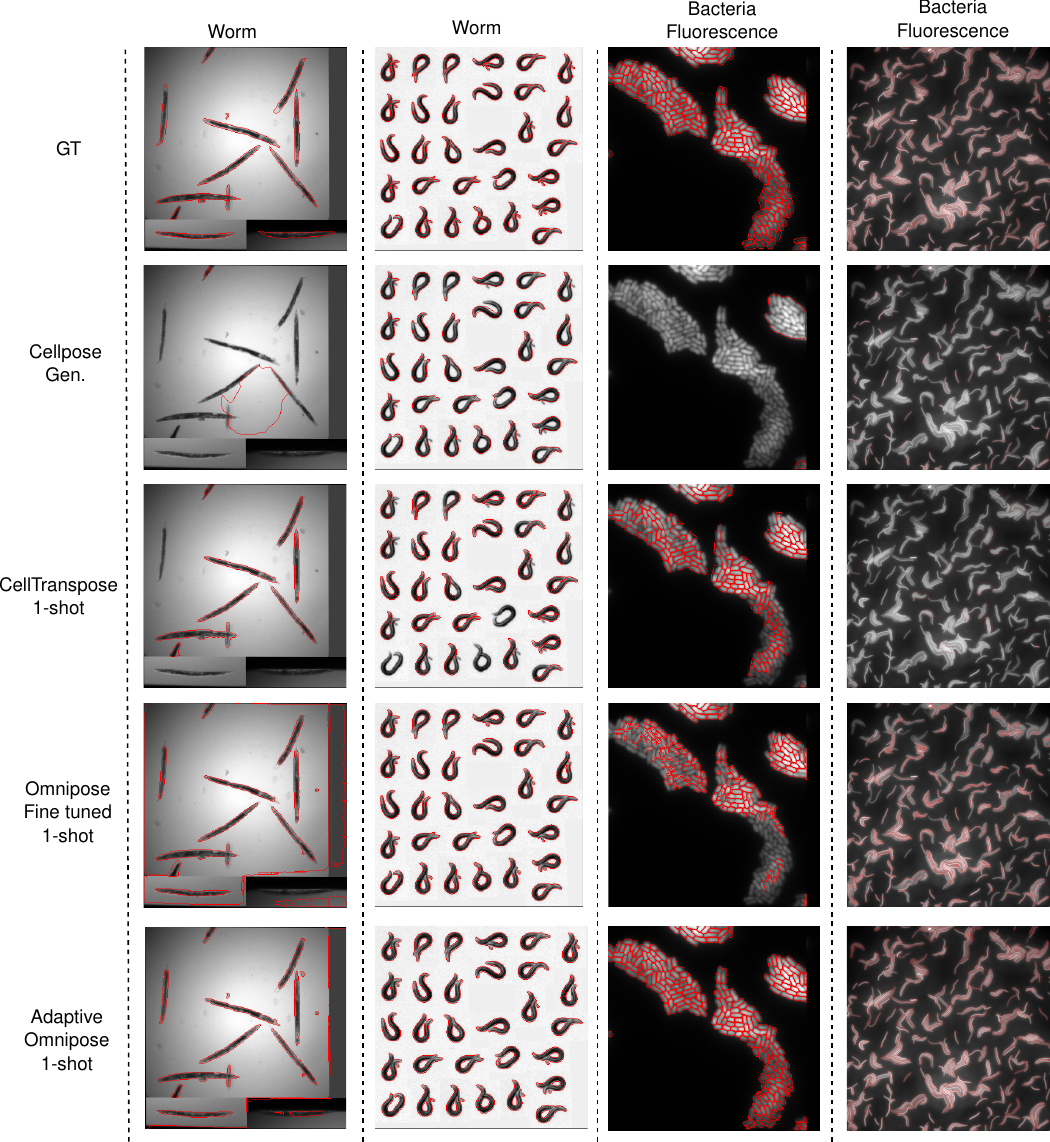}
\caption{\textbf{Qualitative results.} Samples from the Worm (left two columns) and Bacteria Fluorescence (right two columns) datasets highlighting the variability of cell morphologies, and the corresponding segmentations by different approaches. GT indicates ground-truth.}
\label{fig:morphology_samples}
\vspace{-5mm}
\end{figure}

\section{Experiments}
\label{sec:experiments}

\textbf{Implementation.}
We evaluate our approach, named \emph{Adaptive Omnipose}, with different number of shots, $K = 1, 2, 3, 5, 10$. We calculate the average precision (AP) at an intersection over union (IoU) of 0.5. As source datasets $\mathcal{D}^s$ we use separately Bacteria Phase and Bacteria Fluorescence, which are the largest and with the most diverse morphologies among those released by~\cite{Cutler2022-js}.

\begin{table*}[t!]
\begin{center}
\caption{AP results at IoU 0.5: Adapting from Bacteria Phase (BP) to Bacteria Fluorescence (BF) and Worm and from Bacteria Fluorescence  (BF) to Bacteria Phase (BP) and Worm.}
\label{table:adapt}
\scalebox{0.55}{
\begin{tabular}{l|lllll|lllll|lllll|lllll}
\hline\noalign{\smallskip}
{$\mathcal{D}^s$ $\rightarrow$ $\mathcal{D}^t$} & \multicolumn{5}{c}{BP $\rightarrow$ BF} & \multicolumn{5}{c}{BP $\rightarrow$ Worm} & \multicolumn{5}{c}{BF $\rightarrow$ BP} & \multicolumn{5}{c}{BF $\rightarrow$ Worm}\\
& 1-shot & 2-shot & 3-shot & 5-shot & 10-shot & 1-shot & 2-shot & 3-shot & 5-shot & 10-shot & 1-shot & 2-shot & 3-shot & 5-shot & 10-shot & 1-shot & 2-shot & 3-shot & 5-shot & 10-shot \\

\hline		
{\bf Cellpose-Gen} 					    & 0.137 & 0.137 & 0.137 & 0.137 & 0.137 & 0.173 & 0.173 & 0.173 & 0.173 & 0.173 & 0.335 & 0.335 & 0.335 & 0.335 & 0.335 & 0.173 & 0.173 & 0.173 & 0.173 & 0.173 \\
{\bf Omnipose-LB} 					    & 0.009 & 0.009 & 0.009 & 0.009 & 0.009 & 0.470 & 0.470 & 0.470 & 0.470 & 0.470 & 0.001 & 0.001 & 0.001 & 0.001 & 0.001 & 0.002 & 0.002 & 0.002 & 0.002 & 0.002 \\
{\bf Omnipose-FT} 			            & 0.431 & 0.511 & 0.504 & 0.512 & 0.578 & 0.515 & 0.501 & 0.511 & 0.517 & 0.554 & 0.238 & 0.364 & 0.384 & 0.375 & 0.446 & 0.483 & 0.501 & 0.500 & 0.511 & 0.547 \\
{\bf Omnipose-UB} 					    & 0.920 & 0.920 & 0.920 & 0.920 & 0.920 & 0.847 & 0.847 & 0.847 & 0.847 & 0.847 & 0.889 & 0.889 & 0.889 & 0.889 & 0.889 & 0.847 & 0.847 & 0.847 & 0.847 & 0.847 \\
{\bf Cellpose-UB} 					    & 0.927 & 0.927 & 0.927 & 0.927 & 0.927 & 0.792 & 0.792 & 0.792 & 0.792 & 0.792 & 0.717 & 0.717 & 0.717 & 0.717 & 0.717 & 0.792 & 0.792 & 0.792 & 0.792 & 0.792 \\
{\bf Stardist-UB} 					    & 0.376 & 0.376 & 0.376 & 0.376 & 0.376 & 0.326 & 0.326 & 0.326 & 0.326 & 0.326 & 0.341 & 0.341 & 0.341 & 0.341 & 0.341 & 0.326 & 0.326 & 0.326 & 0.326 & 0.326 \\
{\bf CellTranspose} 					& 0.375 & 0.493 & 0.516 & 0.529 & 0.613 & 0.531 & 0.595 & 0.622 & 0.645 & 0.650 & 0.142 & 0.181 & 0.263 & 0.359 & 0.403 & 0.486 & 0.539 & 0.520 & 0.548 & 0.564 \\
\hline
{\bf Adaptive Omnipose }		        & 0.594 & 0.601 & 0.603 & 0.623 & 0.642 & 0.592 & 0.638 & 0.642 & 0.645 & 0.656 & 0.409 & 0.428 & 0.447 & 0.461 & 0.471 & 0.557 & 0.567 & 0.581 & 0.596 & 0.585 \\
\hline
\end{tabular}
}\end{center}
\end{table*}

\begin{table}
\begin{center}
\caption{Ablation on 1-shot adaptation. AP at 0.5 IoU.}
\label{table:ablation}
\scalebox{0.65}{
\begin{tabular}{l|c|c}
\hline
{\bf }                                                                                  & BP $\rightarrow$ BF & BP $\rightarrow$ Worm\\
\hline
{\bf Omnipose-UB }	                                                                    & 0.920 & 0.847 \\			
{\bf Adaptive Omnipose}	                                                                & 0.594 & 0.592 \\
{\bf No $\boldsymbol{\mathcal{L}^{CB}}$}	                                            & 0.525 & 0.561 \\
{\bf No $\boldsymbol{\mathcal{L}^{CD}}$}	                                            & 0.569 & 0.553 \\
{\bf No $\boldsymbol{\mathcal{L}^{CB}}$ \&  No $\boldsymbol{\mathcal{L}^{CD}}$ }	    & 0.431 & 0.515 \\
\hline
\end{tabular}
}\end{center}
\end{table}
We pretrain our model for 3800 epochs with constant learning rate of 0.03, and the RAdam optimizer with weight decay ${10}^{-5}$ and batch size 4. Adaptation is done by following the data splitting guidelines from~\cite{Keaton2023}.
We randomly sample $K$ shot samples from the respective training set of the target dataset $\mathcal{D}^t$. Hyperparameters for the adaptation losses include $|\mathcal{N}_i|=20$, $\tau$ = 0.1, $m$ = 10, $\gamma_1$ = 0.05, and $\gamma_2$ = 0.05. We used an image patch size of $h = w = 112$ without rescaling since most cellular metrics such as mean diameter and area were consistent among all dataset groups during training. However, we rescale target images using the ratio of mean diameter across the source dataset and mean diameter across the target. Also, we note that the original Omnipose~\cite{Cutler2022-js} uses a patch size of $224\times 224$, whereas our implementation of all the algorithms uses $112\times 112$. The adaptation procedure takes approximately 3 minutes to complete using an NVIDIA RTX 3090 GPU.

{\bf Bacteria Phase (BP). } 
BP~\cite{Cutler2022-js} is a large bacterial phase-contrast dataset consisting of bacterial mutations and treatments causing extreme morphological variations. Also, it includes various wild-type and mutant bacterial growth under cephalexin and aztreonam. It has a total of 27,500 cells in the training set, and 19,500 in the test set. We use the same split as in the original paper to pretrain the base model.  

{\bf Bacteria Fluorescence (BF).}
The BF~\cite{Cutler2022-js} dataset involves images with cytosol and membrane fluorescence, which has a different modality compared to BP. It consists of 33,200 cells split in train (18,613 cells) and test (14,587 cells) sets.

{\bf Worm.}
Worm~\cite{Cutler2022-js} is a non-bacterial dataset consisting of 1,279 worms in the train set and 1,264 worms in the test set. It was acquired from the Open Worm Movement database~\cite{javer2018open}, and the dead C. elegans from the BBBC010 dataset~\cite{ljosa2012annotated}.

\textbf{Compared methods.}
Cellpose-Gen is Cellpose trained on the generalist dataset of~\cite{Stringer2021-yw}. Omnipose-LB is Omnipose trained on $\mathcal{D}^s$ (BP or BF) and represents a lower bound. Omnipose-FT is Omnipose trained on $\mathcal{D}^s$ and fine-tuned on $\mathcal{D}_K^t$, i.e., with $K$-shot samples, data augmentation and slow learning rate. Stardist-UB~\cite{schmidt2018}, Cellpose-UB and Omnipose-UB are trained and tested on the respective training and testing splits of the target datasets and represent upper bounds. CellTranspose~\cite{Keaton2023} and Adaptive Omnipose are pretrained on $\mathcal{D}^s$ and then adapted with $\mathcal{D}_K^t$. See Figure~\ref{fig:morphology_samples} for qualitative results.

{\bf BP $\rightarrow$ BF and BF $\rightarrow$ BP.} The modality difference between BP and BF adds to the inherent distribution shift: BF images look almost as negative images of BP~\cite{Cutler2022-js}. This is reflected in Table~\ref{table:adapt}, where Omnipose-LB shows catastrophic performance degradation. Adaptive Omnipose steadily improves, and is very effective even with 1-shot. Also, as expected Adaptive Omnipose consistently surpasses CellTranspose, since CellTranspose builds on Cellpose, which was not designed to handle morphologies where a cell center is not clearly defined.

{\bf  BP $\rightarrow$ Worm and BF $\rightarrow$ Worm.} The target dataset is Worm. The trend is similar to the BP $\rightarrow$ BF and BF $\rightarrow$ BP cases, where Adaptive Omnipose is effective even with 1-shot and consistently improves, and is performing above Omnipose-FT and CellTranspose.

\textbf{Ablation Study.}
Since Adaptive Omnipose adds two new losses to the Omnipose framework, we test it by removing one loss at a time on a 1-shot adaptation scheme. Table~\ref{table:ablation} shows that including even one loss leads to significant improvement over just fine-tuning the model, while adding both leads to the best improvement. 


\section{Conclusions}

We introduced a few-shot domain adaptation procedure for cell instance
segmentation. It improves upon prior work because the model it adapts
can handle elongated and non-convex morphologies by design. The
procedure requires very limited manual annotation efforts to be
effective (even one to five cells is enough), and it requires about 3
minutes with an off-the-shelf GPU. Our experiments based on
pretraining on two different sources, and adapting on three different
challenging targets fully support our modeling framework, including a
favorable comparison with previous work.

\section{Compliance with ethical standards}
\label{sec:ethics}

Ethical approval was not required as confirmed by the license attached with the open access data.

\section{Acknowledgments}
\label{sec:acknowledgments}

Research reported in this publication was supported by the National
Institute Of Mental Health of the National Institutes of Health under
Award Number R44MH125238. The content is solely the responsibility of
the authors and does not necessarily represent the official views of the
NIH. This material is also based upon work
supported by the National Science Foundation under Grants No. 1920920, 2223793.

\bibliographystyle{IEEEbib}
{\small
  \bibliography{references}
  }

\end{document}